\documentclass[letterpaper, 10 pt, conference]{ieeeconf}  

\usepackage{color}
\IEEEoverridecommandlockouts                              

\usepackage{graphics} 
\usepackage{epsfig} 
\usepackage{amsbsy}
\usepackage{dsfont}
\usepackage{mathtools}
\usepackage{bm}
\usepackage{subfiles}
\usepackage[noend,ruled,linesnumbered]{algorithm2e}
\usepackage{caption}
\usepackage{subcaption}
\usepackage{cite}
\usepackage{amsmath, amssymb}
\usepackage{bbm}

\DeclareCaptionFont{mysize}{\fontsize{8}{9.6}\selectfont}
\newtheorem{remark}{\textbf{Remark}}

\newtheorem{prop}{\textbf{Proposition}}

\newcommand{\mb}[1]		{\mathbf{#1}}

\graphicspath{figures}

\title{\LARGE \bf
Graph Neural Networks for Multi-Robot Active Information Acquisition
}

\author{Mariliza Tzes$^{*}$, Nikolaos Bousias$^{*}$, Evangelos Chatzipantazis and George J. Pappas
\thanks{
The authors are with the GRASP Lab, University of Pennsylvania, Philadelphia, PA 19104, USA, {\tt\small \{mtzes, nbousias, vaghat, pappasg\}@seas.upenn.edu}.}%
\thanks{This work was supported by the ARL grant DCIST CRA W911NF-17-2-0181.}
\thanks{*equal contribution}
}

\begin{document}

\maketitle
\thispagestyle{empty}
\pagestyle{empty}


\begin{abstract}
This paper addresses the Multi-Robot Active Information Acquisition (AIA) problem, where a team of mobile robots, communicating through an underlying graph, estimates a hidden state expressing a phenomenon of interest. Applications like target tracking, coverage and SLAM can be expressed in this framework. Existing approaches, though, are either not scalable, unable to handle dynamic phenomena or not robust to changes in the communication graph. To counter these shortcomings, we propose an Information-aware Graph Block Network (I-GBNet), an AIA adaptation of Graph Neural Networks, that aggregates information over the graph representation and provides sequential-decision making in a distributed manner. The I-GBNet, trained via imitation learning with a centralized sampling-based expert solver, exhibits permutation equivariance and time invariance, while harnessing the superior scalability, robustness and generalizability to previously unseen environments and robot configurations. Experiments on significantly larger graphs and dimensionality of the hidden state and more complex environments than those seen in training validate the properties of the proposed architecture and its efficacy in the application of localization and tracking of dynamic targets.

\end{abstract}


\section{Introduction}

Over recent years, 
we have seen UAVs participating in the fighting of wildfires in California, robotic snakes assisting in localization of earthquake victims or UUVs deployed for underwater ocean mapping \cite{rubio2019review}. 
In such scenarios, autonomous robots need to sequentially make decisions that would allow them to 
better estimate the extent of the fire and decide upon extinguishing locations. The process of designing efficient paths to actively estimate a hidden state that expresses such a phenomenon of interest by utilizing measurement readings of on-board sensors is known in the literature as \textit{Active Information Acquisition} (AIA). An exhaustive list of applications can be formulated in the AIA setup, from environmental monitoring, search \& rescue, surveillance and coverage, target tracking and localization and active-SLAM
 \cite{ma2017informative,frew2008target, responders, bousias2019collaborative, bucci2019decentralized, carlone2014active, roy_slam, DBLP:journals/corr/abs-1910-10754}.
 
 A plethora of existing works \cite{chung2006decentralized, hoffmann2009mobile, guestrin2005near, dames2012decentralized, le2009trajectory, singh2009efficient, schlotfeldt2018anytime, atanasov2015decentralized, levine2010information, hollinger2014sampling, kantaros2019asymptotically, best2019dec, tzes2021distributed} are proposed that solve the AIA problem. Myopic approaches \cite{chung2006decentralized, hoffmann2009mobile, guestrin2005near, dames2012decentralized} rely on computing controllers that incur the maximum immediate decrease of an uncertainty measure while search-based nonmyopic schemes \cite{le2009trajectory, singh2009efficient, schlotfeldt2018anytime, atanasov2015decentralized} solve the problem by pruning the exploration process. The latter return suboptimal solutions while their decentralized counterparts rely on coordinate descent, making them computationally intractable as the planning horizon and/or the number of robots increases.  Nonmyopic sampling-based approaches \cite{levine2010information, hollinger2014sampling, best2019dec, kantaros2019asymptotically, tzes2021distributed} gained popularity due to their ability to compute informative paths fast.  All these approaches suffer from the following limitations:
 1) they do not \textit{scale}, e.g. for more than dozens of robots, 2) are not \textit{robust} to changes of the problem's parameters, e.g. number of agents or changes to the connectivity of robots and/or 3) do not address dynamic phenomena of interest.
 
 \begin{figure}[t]
     \centering
     \includegraphics[width=0.4\textwidth]{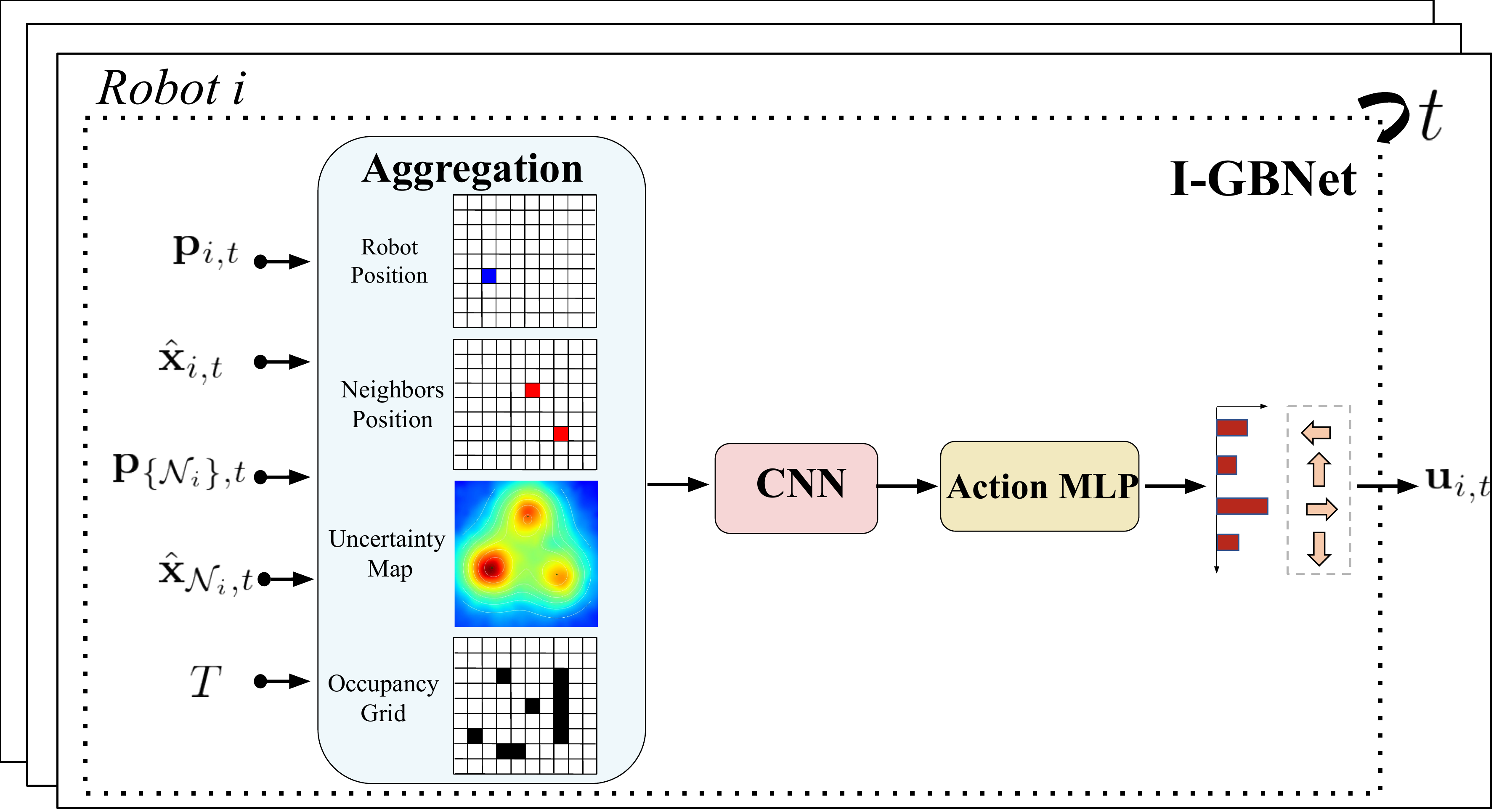}
     \setlength{\belowcaptionskip}{-22 pt}
     \caption{In this Figure we illustrate our graph parametrization of the AIA problem with the I-GBNet. Robot $i$ collects its neighbors' positions and estimates, along with its own attributes and map and feeds them to the network. The latter produces the next action, the robot updates its position takes new measurements. The process is repeated per robot and timestep.}
     \label{fig:intro}
 \end{figure}
 
 Graph Neural Networks (GNNs) are information processing architectures for signals on graphs. Thanks to their distributed structure, GNNs can easily be adapted to multi-robot applications where the nodes and edges represent the robots and communication links respectively. They follow a neighborhood aggregation scheme 
  and are therefore inherently suitable for distributed implementations. 
 GNNs are also promising for their transferability across different graphs and can be designed such that they generalize to previously unseen multi-robot scenarios. Their use has the potential to radically increase (a) robustness to changes in the topology of the graph and (b) scalability with respect to the number of robots.
 
In this paper, we propose a novel architecture based on GNNs that solves the AIA problem in a distributed manner and addresses the  aforementioned challenges. 
Specifically, we translate the multi-robot information gathering problem to a graph representation and formulate it as a sequential decision-making problem, where an Information-aware Graph Block Network (I-GBNet) learns to derive control actions that drive the robots to actively estimate the state of the phenomenon of interest. At every timestep, each robot collects positions and estimates from its neighbors and feeds them along with its own attributes to the I-GBNet that produces the next control inputs to be applied; see also Fig.\ref{fig:intro}. To train our network, we use imitation learning where the expert is a centralized sampling-based algorithm. For this reason, we created a dataset that contains frames of episodes on randomly generated environments with fixed dimensions and a fixed number of robots and dimensionality of the state across all episodes.

\par\textbf{Literature Review:}
An exhaustive literature has been developed that addresses the AIA problem. 
A centralized sampling based approach is introduced in \cite{kantaros2019asymptotically} that explores the joint physical and information space of the robots. In \cite{best2019dec} the authors propose a decentralized Monte Carlo Tree Search \cite{browne2012survey} method that samples the individual action space of each robot and then creates a sparse approximation of the joint action space. 
In \cite{tzes2021distributed}, the authors  design a distributed sampling-based approach where robots build their own trees and share with their neighbors information of randomly sampled nodes of their trees. The authors in \cite{kantaros2021scalable} propose a more scalable approach by tessellating the environment into Voronoi diagrams, but an all-to-all communication and static targets are assumed.
Data-driven methods \cite{choudhury2017learning, he2016active, ruckin2022adaptive, notomista2022multi} have also been proposed to allow for online implementations. Prior work in data-driven information acquisition utilizes imitation learning via a clairvoyant oracle \cite{choudhury2017learning}, \cite{he2016active}. In \cite{ruckin2022adaptive} the authors develop a deep reinforcement learning technique that combines tree search with an
offline-learned neural network, however it is only applicable to a single robot.
To the best of our knowledge, GNNs have never been used before in the AIA setting. Noticeably, they recently started gaining popularity in multi-robot applications \cite{li2020graph, li2021message, gosrich2022coverage, tolstaya2020learning, tolstaya2021multi}. In \cite{li2021message} GNNs are employed for path planning where an attention mechanism prioritizes important information. The authors in \cite{tolstaya2021multi} propose a GNN based method for coverage. Common in these works is the absence of notion of noisy observations of the robots' surroundings. 

\par\textbf{Contributions:} This paper proposes the \textit{first} method for Active Information Acquisition using Graph Neural Networks. We designed a network that is \textit{scalable} with respect to the number of robots and dimensionality of the hidden state to be estimated and \textit{generalizes} to previously unseen multi-robot configurations. Our method is \textit{robust} to communication failures and can be applied for \textit{time-varying} communication graphs and \textit{dynamic} hidden state. Additionally, we provide a variety of \textit{quantitative} and \textit{qualitative} experiments that illustrate the efficacy of our proposed architecture for the example of target localization and tracking.
\section{Problem Formulation}
\label{sec:problem_formulation}
Consider a homogeneous team of $N$ mobile robots that reside in a complex environment $T \subset \mathbb{R}^{2}$. The environment is cluttered with a set of arbitrarily shaped obstacles $\mathcal{O} \subset T$, thus forming the obstacle-free area $T_{\text{free}} := T \setminus \mathcal{O}$. The robots are governed by the \textit{robot dynamics}
    $\mb{p}_{i,t+1} = f(\mb{p}_{i,t},\mb{u}_{i,t}), \ \forall i \in \{1,\dots,N\}$
where $\mb{p}_{i,t} \in T_{\text{free}}$ and $\mb{u}_{i,t} \in \mathcal{U}$ denote the state of robot $i$ at time $t$ and control action selected from a finite set of admissible control actions, $\mathcal{U}$. 
The robots are tasked with collaboratively estimating a hidden state $\mb{x}_t \in \mathbb{R}^{d_x}$ with \textit{hidden state dynamics}:
\vspace{-1mm}
\begin{align}
    \mb{x}_{t+1} = A \mb{x}_t + \mb{w}_t, \quad \mb{w}_t \sim \mathcal{N}(0, \mb{Q}_t)
\label{eq:hidden_dynamics}
\end{align}
where $\mathbf{Q}_t$ is the noise covariance matrix. The robots are equipped with sensors capable of collecting noisy measurements of the hidden state as per the \textit{observation model}
\vspace{-1mm}
\begin{align}
    \mb{y}_{i,t} = M(\mb{p}_{i,t})\mb{x}_t + \mb{v}_{i,t}, \quad \mb{v}_{i,t} \sim \mathcal{N}(0, \mb{R}_{i})
\label{eq:observation_model}
\end{align}
where $\mb{y}_{i,t}$ is the measurement of robot $i$ at time $t$. We assume that the transition matrices $\{A,M\}$ and the noise covariance matrices $\{\mb{Q}_t, \mb{R}_i \}$ are known, similar to \cite{tzes2021distributed} and the robots perfectly localize themselves in a global frame. Hereafter, when subscript $i$ is omitted we denote compactly attributes of all robots.

\subsection{Multi-Robot Active Information Acquisition}
\label{subsec:active_information_acquisition}
The AIA problem requires the robots to collaboratively estimate the hidden state $\mathbf{x}_t$. To do so, the robots maintain a prior Gaussian distribution for the hidden state, $\hat{\mb{x}}_0 \sim \mathcal{N}(\bm{\mu}_0, \Sigma_0)$. Given the prior distribution and measurements received up to time $t$, $\mathbf{y}_{0:t}$,  the \`{a}-posteriori distribution is computed i.e., $\hat{\mb{x}}_t \sim \mathcal{N}(\bm{\mu}_t(\mb{y}_{0:t}), \Sigma_t(\mb{y}_{0:t}))$ where $\bm{\mu}_t(\mb{y}_{0:t})$ and $\Sigma_t(\mb{y}_{0:t})$ denote the \`{a}-posteriori mean and covariance matrix respectively. An indicative function of the overall quality of the measurements up to time $t$ in estimating $\mb{x}_t$ is the determinant of the \`{a}-posteriori covariance matrix, i.e $\det \Sigma_t(\mb{y}_{0:t})$ \cite{atanasov2014joint}. Alternative information measures could be used such as the trace or the maximum eigenvalue of the covariance. Given initial robots' positions $\mb{p}_0$ and prior distribution $\hat{\mb{x}}_0$, the goal is to compute a planning horizon $F$ and a sequence of control inputs $\bm{u}_{0:F}$, which solves the following deterministic optimal control problem:
\vspace{-1mm}
\begin{subequations}
\label{eq:Prob1}
\begin{align}
& \min_{\substack{ F, \bm{u}_{0:F}}} \left[J(F,\bm{u}_{0:F}) = \sum_{t=0}^{F}  \det\Sigma_{t+1} \right] \label{eq:obj1}\\
& \ \ \ \ \ \ \ \det\Sigma_{F+1}\leq \epsilon \label{eq:constr11}, \\ 
& \ \ \ \ \ \ \ \    \mb{p}_{t+1} \in T_{\text{free}}^N, \label{eq:constr12} \\
& \ \ \ \ \ \ \ \  \mb{p}_{i,t+1} = f(\mb{p}_{i,t},\mb{u}_{i,t}), \quad \forall i=\{1,\dots, N\} \label{eq:constr13}\\
& \ \ \ \ \ \ \ \ \Sigma_{t+1} =\rho(\mb{p}_{t},\Sigma_t\label{eq:constr14})
\end{align}
\end{subequations}
where the objective \eqref{eq:obj1} captures the cumulative uncertainty in the estimation of $\mb{x}(t)$ after fusing information collected by all robots from $t=0$ up to time $F$ and the constraint in \eqref{eq:constr11} requires that the final uncertainty of the state $\mb{x}(F)$ is less than a user-specified threshold $\epsilon$.  In \eqref{eq:constr12} we require obstacle free paths and in \eqref{eq:constr14}, $\rho(\cdot)$ stands for the Kalman Filter update rule that computes the \`{a}-posteriori distribution. A Centralized Sampling-Based (C-SB) method that solves Problem \eqref{eq:Prob1} is introduced in \cite[Section III]{kantaros2019asymptotically} that is probabilistically complete and asymptotically optimal.

 \begin{remark}
 \label{ref:separation_principle}
     The optimization problem defined in \eqref{eq:Prob1} resulted from a stochastic optimal control problem defined in \cite[Section III]{tzes2021distributed} where a separation principle \cite{atanasov2014information} is applied due to the linear Gaussian assumptions.
\end{remark}

\subsection{Problem Statement}
Given initial positions of the robots $\mathbf{p}_0$, capable of taking measurements as per the observation model \eqref{eq:observation_model}, communicating via an underlying communication network and prior estimates $\hat{\mb{x}}_0$, we design a GNN architecture that (i) solves \eqref{eq:Prob1} in a distributed manner (ii) is trained using imitiation learning to mimic the C-SB expert and (iii) derives sequential control actions $\mathbf{u}_{t}$ that allow the robots to actively decrease their uncertainty over $\mathbf{x}_t$.

\section{Graph Neural Networks for AIA}
\begin{figure*}[h!]
    \captionsetup{font=footnotesize}
    \centering
    \includegraphics[width=\textwidth]{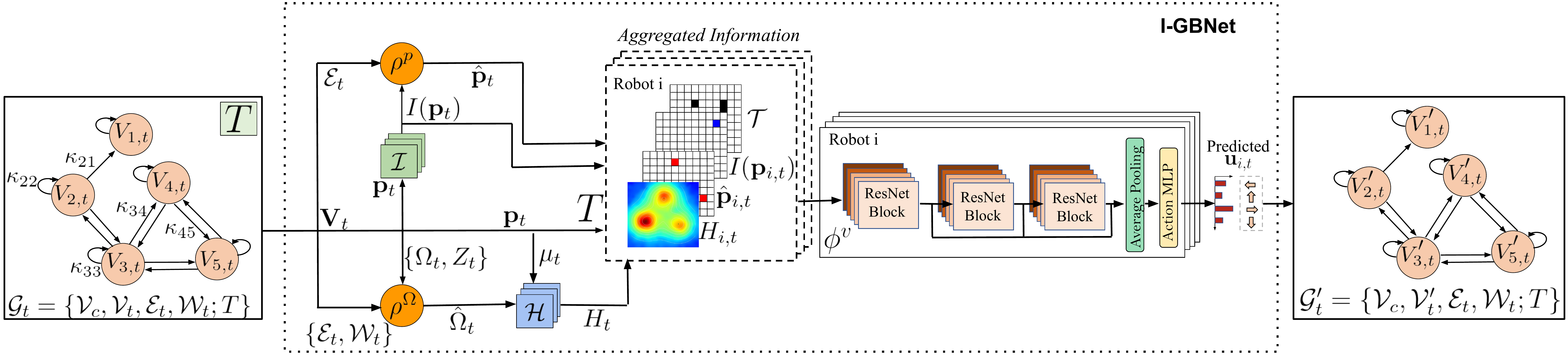}
    \setlength{\belowcaptionskip}{-18 pt}
    \setlength{\abovecaptionskip}{-8 pt}
    \caption{I-GBNet structure comprising of aggregation functions $\rho^\Omega,\rho^p$, a heatmap projection module $\mathcal{H}$ that projects the posterior distribution of the hidden state on the grid, and a node update function that produces admissible control actions. It only updates the node attributes of the graph, leaving the latter topologically unscathed. The update node function $\phi^v$ comprises of 3 consecutive ResNet Blocks with skip connections, an adaptive average pooling layer and an Action MLP.}
    \label{fig:Graph Block}
\end{figure*} 
To solve the Multi-Robot Active Information Acquisition problem described in Section \ref{sec:problem_formulation} in a distributed way with GNNs, we develop a suitable graph representation of the problem. At every timestep the graph is given as an input to the network that consists of an Information-aware Graph Block Network (I-GBNet) and a Multi-Layer Perceptron (MLP). The I-GBNet updates the node attributes of the incoming graph and the updated attributes are fed to the MLP that derives the control actions for all robots. 

\subsection{Graph Representation of AIA problem}
\label{subsec:graph_representation}


In the distributed setting, we assume that each robot maintains a local estimate over the hidden state expressed via its \`{a}-posteriori Gaussian distribution $\hat{\mb{x}}_{i,t} \sim \mathcal{N}(\bm{\mu}_{i,t}, \Sigma_{i,t})$. For simplicity reasons, we define $Z_{i,t} := M(\mb{p}_{i,t})^T \mb{R}_{i,t}^{-1} M(\mb{p}_{i,t})$ and the information matrix $ \Omega_{i,t} = \Sigma_{i,t}^{-1}$.

Let the robots communicate through an underlying communication network modeled as a directed \textit{communication graph} $\mathcal{G}_{c,t}=(\mathcal{V}_{c},\mathcal{E}_{c,t})$. The set of vertices $\mathcal{V}_{c} \coloneqq \{1,\dots,N\}$ are indexed by the robots and an edge $(i,j) \in \mathcal{E}_{c,t} \subseteq \mathcal{V}_{c} \times \mathcal{V}_{c}$ defines a direct communication between robots $i$ and $j$. Robot $i$ can exchange information with the set of its neighbors $\mathcal{N}_{i,t}$. Hereafter we define $\tilde{\mathcal{N}}_{i,t} \coloneqq \mathcal{N}_{i,t} \cup \{i\}$.


Given the aforementioned communication graph $\mathcal{G}_{c,t}$, let the state of the problem at timestep $t$ be described by the following homogeneous, weighted, directed \textit{network graph}: 
\vspace{-1mm}
\begin{equation}
    \mathcal{G}_t=\{\mathcal{V}_c, \mathcal{V}_t, \mathcal{E}_t , \mathcal{W}_t;\mathbf{g}\}
    \label{eq:G_t}
\end{equation}
where $\mathcal{V}_t=\{\mathbf{V}_{i,t}\}_{i=1}^N$ is the set of node attributes of cardinality $\vert \mathcal{V}_t \vert=N$, indexed by the robots $i \in \mathcal{V}_c$, 
$\mathcal{E}_t=\mathcal{E}_{c,t} \cup \{(i,i), \forall i\in \mathcal{V}_c \}$ is the set of edges, $\mathcal{W}_t \in \mathbb{R}^{N \times N}$ is the weight matrix and $\mb{g}$ represents the global attributes of the graph (e.g. environment $T$). Each node (robot) $i$ contains the following \textit{node attribute}:
\vspace{-1mm}
\begin{equation}
    \mb{V}_{i,t}= \{ \mb{p}_{i,t}, Z_{i,t}, \bm{\mu}_{i,t}, \Omega_{i,t} \}
    \label{eq:node_attribute}
\end{equation}
\vspace{-0.3mm}
The weight matrix quantifies the importance of information being propagated from robot $i$ to $j$ and is defined as $[\mathcal{W}_{t}]_{ij}= \kappa_{ij} \mathds{1}_{(i,j) \in \mathcal{E}_t}$,  where $\kappa_{ij} > 0$ and $\sum_{j \in \tilde{\mathcal{N}}_{i,t}} \kappa_{ij} = 1$. 

\subsection{Information-aware Graph Block Network}\label{subsection:Graph_Block}
The core of our GNN is the Information-aware Graph Block Network (I-GBNet) $\Phi$, illustrated in Fig.\ref{fig:Graph Block} and inspired by \cite{DBLP:journals/corr/abs-1806-01261}. I-GBNet receives a graph signal $\mathcal{G}_t$ and updates the node attributes by propagating information through the graph, while keeping the graph topology unchanged, i.e. $\mathcal{G}^{\prime}_t = \Phi(\mathcal{G}_t)$. For the design of I-GBNet, we finely discretize the environment $T \subset \mathbb{R}^2$ and represent it via an occupancy grid-map $\mathcal{T}$ of size $\mathcal{T}_H \times \mathcal{T}_W$. The network $\Phi$ comprises of three parts i) two aggregation functions $\rho^{\Omega}(\cdot), \rho^p(\cdot)$, (ii) a heatmap projection function $\mathcal{H}$ and (iii) a node-update function $\phi^v$. 

\textbf{Aggregation}: We assume that at timestep $t$, the robot $i$ receives information of the form $\{(\Omega_{j,t}, \mathcal{I}(\mb{p}_{j,t})), \forall j \in \tilde{\mathcal{N}}_{i,t} \}$, where $\mathcal{I}(\cdot)$ is a function that transforms the robot's position into a binary robot-grid map. Let $\mathbf{\Omega}_{\tilde{\mathcal{N}}_{i,t},t}=\{\Omega_{j,t}, \forall j \in \tilde{\mathcal{N}}_{i,t} \}$ and $\mathbf{I}_{\tilde{\mathcal{N}}_{i,t},t}= \{\mathcal{I}(\mb{p}_{j,t}), \forall j \in \tilde{\mathcal{N}}_{i,t} \}$. Inspired by the active information problem setup and constraint \eqref{eq:constr14}, the aggregation function $\rho^{\Omega}$ is defined as the Distributed Kalman Filter (DKF), proposed in \cite{atanasov2014joint}, where robot $i$ updates its \`{a}-posteriori covariance matrix as follows:
\vspace{-2mm}
    \begin{equation}
    \hat{\Omega}_{i,t} = \rho^{\Omega}(\mathbf{\Omega}_{\tilde{\mathcal{N}}_{i,t},t}, Z_{i,t}; \mathcal{W}_t) = \sum\limits_{j \in \tilde{\mathcal{N}}_{i,t}} \kappa_{ij} \Omega_{j,t} + Z_{i,t} \label{eq:rho_omega}
    \end{equation}

Each robot, also, aggregates the positions of its neighbours into a neighbors-grid map as:
\begin{align}
    \hat{\mathbf{p}}_{i,t} = \rho^p( \mathbf{I}_{\tilde{\mathcal{N}}_{i,t},t}) = \sum_{j \in \tilde{\mathcal{N}}_{i,t}} \mathcal{I}(\mathbf{p}_{j,t}) - \mathcal{I}(\mathbf{p}_{i,t}) \label{eq:rho_p}
\end{align}

The selected aggregation functions are invariant to permutations of their inputs by construction.

\textbf{Heatmap Projection}: Once the robot computes the aggregated covariance matrix $\hat{\Omega}_{i,t} \in \mathbb{R}^{d_x \times d_x}$, it projects it into the occupancy grid map $\mathcal{T}$ via the projection function $\mathcal{H}\colon \{\hat{\Omega}_{i,t}, \bm{\mu}_{i,t}\} \rightarrow H_{i,t}$, where $H_{i,t}$ is the resulting grid heatmap that represents the magnitude of uncertainty in $\mathcal{T}$. In the case of multi-target localization, the projection module would produce a combination of the posterior Gaussian heatmaps.

\textbf{Node-Update}: Given the aggregated neighbors-grid map $\hat{\mathbf{p}}_{i,t}$, the robot-grid map $\mathcal{I}(\mb{p}_{i,t})$, the heatmap-grid $H_{i,t}$ and the occupancy-grid map $\mathcal{T}$, the node-update function $\phi^v$ produces the control actions
\vspace{-1mm}
\begin{equation}
    \mb{u}_{i,t} \coloneqq \phi^v(\hat{\mathbf{p}}_{i,t}, \mathcal{I}(\mb{p}_{i,t}), H_{i,t}; \mathcal{T})
    \label{eq:phi_v}
    \vspace{-1mm}
\end{equation}
hence the updated node attributes are $\mathbf{V}_{i,t}^{\prime} = \{\mb{u}_{i,t} \}$.
The node-update function $\phi^v$ is parametrized by a Convolutional Neural Network with learnable parameters $\theta_{\text{CNN}}$ composed with an Action-MLP of parameters $\theta_{\text{MLP}}$.
The output of the Action MLP produces a categorical distribution over the set of admissible control inputs $\mathcal{U}$ from which the most probable action $\mb{u}_{i,t}$ is selected. 

\subsection{Properties of Graph Block Network}

The graph topology is expected to vary over time owing to the mobility of the agents.
Therefore, the Graph Block is required to perform consistently for all permutations of $\tilde{\mathcal{N}}_{i,t} \subseteq\mathcal{V}_c$, i.e. swapped order of robot indices, and regardless of the time shift. The permutation equivariance property for the aforementioned Graph Block network, directly limits the required sample complexity for training the network. 
\begin{prop}(Permutation equivariance of Graph Block)
\label{Proposition1}
Given the aforementioned graph $\mathcal{G}_t$, for any permutation $\pi:[n]\rightarrow [n]$ of $\mathcal{V}_c$, with corresponding permutation matrix $P_{\pi}$, and the Graph Block defined in subsection \ref{subsection:Graph_Block}, it holds that $\Phi(\pi(\mathcal{G}_t)) = \pi(\Phi(\mathcal{G}_t))$.
\end{prop}
\begin{proof}
For the purposes of the proof, we denote 
$\mathbf{\hat{\Omega}}_t := \oplus_{i\in [N]}\hat{\Omega}^T_{i,t}$, $\mathbf{\Omega_t} := \oplus_{i\in [N]}\mathbf{\Omega}^T_{i,t}$, $\mathbf{Z}_t := \oplus_{i\in [N]}Z^T_{i,t}$, $\hat{\mathbf{P}}_t := \oplus_{i\in [N]}\hat{\mathbf{p}}_{i,t}^T,$ and $\mathbf{I}_t := \oplus_{i\in [N]}\mathcal{I}(\mathbf{p}_{i,t})^T$, where the operator $\oplus_{i}$ denotes a concatenation of matrices along the first dimension. Let $\mathbb{A} \in \mathbb{R}^{N \times N}$ denote the adjacency matrix of the graph $\mathcal{G}_t$.
The per-agent aggregation process, described by \eqref{eq:rho_omega} and \eqref{eq:rho_p}, can be reformulated as
\begingroup
\allowdisplaybreaks
\begin{align*}
    \mathbf{\hat{\Omega}}_t &= \begin{bmatrix} \rho^{\Omega}(\mathbf{\Omega}_{\tilde{\mathcal{N}}_{1,t},t}, Z_{1,t}; \mathcal{W}_t)^T ... \   \rho^{\Omega}(\mathbf{\Omega}_{\tilde{\mathcal{N}}_{N,t},t}, Z_{N,t}; \mathcal{W}_t)^T \end{bmatrix}^T \\
    &= (\mathcal{W}_t \otimes \mathbb{I}_{d_x}) \mathbf{\Omega_t} + \mathbf{Z}_t \\ 
    \hat{\mathbf{P}}_t &= \begin{bmatrix} \rho^p( \mathbf{I}_{\tilde{\mathcal{N}}_{1,t},t})^T ... \ \rho^p( \mathbf{I}_{\tilde{\mathcal{N}}_{N,t},t})^T \end{bmatrix}^T = ((\mathbb{A}-\mathbb{I}_{ N})\otimes\mathbb{I}_{\mathcal{T}_H}) \mathbf{I}_t
\end{align*}
\endgroup
Then, the permutation of the aggregated node attributes is the aggregation of the permuted node attributes as 
\begingroup
\allowdisplaybreaks
\begin{align*}
    &\begin{bmatrix} \rho^{\Omega}(\mathbf{\Omega}_{\tilde{\mathcal{N}}_{\pi(1),t},t}, Z_{\pi(1),t}; \pi(\mathcal{W}_t)) \\ \vdots \\ \rho^{\Omega}(\mathbf{\Omega}_{\tilde{\mathcal{N}}_{\pi(N),t},t}, Z_{\pi(N),t}; \pi(\mathcal{W}_t)) \end{bmatrix}=\\
    &=(P_{\pi}\mathcal{W}_t P_{\pi}^T \otimes \mathbb{I}_{d_x}) (P_{\pi} \otimes \mathbb{I}_{d_x})\mathbf{\Omega_t} +(P_{\pi} \otimes \mathbb{I}_{d_x}) \mathbf{Z}_t \\
    &= (P_{\pi}\mathcal{W}_t P_{\pi}^TP_{\pi} \otimes \mathbb{I}_{d_x}) \mathbf{\Omega_t} +(P_{\pi} \otimes \mathbb{I}_{d_x}) \mathbf{Z}_t \\
    &= (P_{\pi} \otimes \mathbb{I}_{d_x}) (\mathcal{W}_t \otimes \mathbb{I}_{d_x}) \mathbf{\Omega_t} +(P_{\pi} \otimes \mathbb{I}_{d_x}) \mathbf{Z}_t \\
    &=(P_{\pi} \otimes \mathbb{I}_{d_x}) \left( (\mathcal{W}_t \otimes \mathbb{I}_{d_x}) \mathbf{\Omega_t} + \mathbf{Z}_t \right)  =\pi\left( \mathbf{\hat{\Omega}}_t \right)
\end{align*}
\endgroup
and similarly,
\begingroup
\allowdisplaybreaks
\begin{align*}
    &\begin{bmatrix} \rho^p( \mathbf{I}_{\tilde{\mathcal{N}}_{\pi(1),t},t})^T & \dots & \rho^p( \mathbf{I}_{\tilde{\mathcal{N}}_{\pi(N),t},t})^T \end{bmatrix}^T \\
    &= (P_\pi(\mathbb{A}-\mathbb{I}_N)P_\pi^T\otimes\mathbb{I}_{\mathcal{T}_H}) (P_\pi\otimes\mathbb{I}_{\mathcal{T}_H}) \mathbf{I}_t\\
    &= (P_\pi\otimes\mathbb{I}_{\mathcal{T}_H})[((\mathbb{A}-\mathbb{I}_N)\otimes\mathbb{I}_{\mathcal{T}_H})\mathbf{I}_t] = \pi(\hat{\mathbf{P}}_t)
\end{align*}
\endgroup
Since the update function is applied on a per--node basis on the aggregated permuted node attributes 
\begin{align*}
    &\begin{bmatrix} \mb{V}_{\pi(1),t}^{\prime}\\ \vdots \\ \mb{V}_{\pi(N),t}^{\prime}\end{bmatrix} = \begin{bmatrix} \phi^v(\hat{\mathbf{p}}_{\pi(1),t}, \mathcal{I}(\mb{p}_{\pi(1),t}), H_{\pi(1),t}; \mathcal{T}) \\ \vdots \\ \phi^v(\hat{\mathbf{p}}_{\pi(N),t}, \mathcal{I}(\mb{p}_{\pi(N),t}), H_{\pi(N),t}; \mathcal{T}) \end{bmatrix} \\
    &=(P_{\pi} \otimes \mathbb{I}_{u})\begin{bmatrix} \phi^v(\hat{\mathbf{p}}_{1,t}, \mathcal{I}(\mb{p}_{1,t}), H_{1,t}; \mathcal{T}) \\ \vdots \\ \phi^v(\hat{\mathbf{p}}_{N,t}, \mathcal{I}(\mb{p}_{N,t}), H_{N,t}; \mathcal{T}) \end{bmatrix}=\pi\left( \begin{bmatrix} \mb{V}_{1,t}^{\prime}\\ \vdots \\ \mb{V}_{N,t}^{\prime}\end{bmatrix} \right)
\end{align*}
Thus, as the topology of the graph remains unchanged, we conclude $\Phi(\pi(\mathcal{G}_t)) = \pi(\Phi(\mathcal{G}_t))$.
\end{proof}
\begin{prop}(Time invariance of Graph Block)\\
For $t_1,t_2 \in \mathbb{R}^{+},\; t_1\neq t_2$, given $\mathcal{G}_{t_1}=\{\mathcal{V}_c,\mathbf{V}_{t_1}, \mathcal{E}_{t_1} , \mathcal{W}_{t_1};\mathbf{g}\}$, $\mathcal{G}_{t_2}=\{\mathcal{V}_c,\mathbf{V}_{t_2}, \mathcal{E}_{t_2} , \mathcal{W}_{t_2};\mathbf{g}\}$ where $\mathbf{V}_{t_1} \equiv \mathbf{V}_{t_2}, \mathcal{E}_{t_1} \equiv \mathcal{E}_{t_2}, \mathcal{W}_{t_1} \equiv \mathcal{W}_{t_2}$, and the Graph Block $\Phi$ defined in subsection \ref{subsection:Graph_Block}, it holds that $\Phi(\mathcal{G}_{t_1}) = \Phi(\mathcal{G}_{t_2})$.
\end{prop}
\begin{proof}
    This proposition is satisfied naturally owing to the imitation learning method explained in the following Section and the fact that none of the components in $\Phi$ is an explicit function of time. The Graph Block network is trained to enable each robot to predict actions consistently and independently of $t$, given a graph $\mathcal{G}_t$, imitating expert decision making. 
\end{proof}
\section{Training and Architecture}

\subsection{Dataset and Training}
\label{sec:dataset_training}
For the dataset creation, we run the C-SB expert for 1500 randomly generated $20m\times 20m$ environments cluttered with a fixed number of obstacles, $10$ robots with first-order dynamics and sensing radius of $2m$, static hidden state $\mb{x} \in \mathbb{R}^{20}$ and $\mathcal{U} = \{left, up,right,down\}$ by $0.5m$. The uncertainty threshold $\epsilon$ was fixed to $1e^{-1}$. For each instantiation we run entire episodes and randomly select 10 timesteps out of each to collect $15000$ sets $\{(\mb{p}^k, Z^k, \bm{\mu}^k, \Omega^k, T^k),  \bm{u}^k\}$. The latter sets are then transformed via the graph representation process (Section \ref{subsec:graph_representation}) and result in our dataset $\mathcal{D}= \{\mathcal{G}^k, \bm{u}^k \}$.
Given dataset $\mathcal{D}$ and the learnable parameters of the I-GBNet, $\theta_{\text{CNN}}$ and $\theta_{\text{MLP}}$, we minimize the cross-entropy loss:
\vspace{-2.5mm}
\begin{equation*}
    \min_{\theta_{\text{CNN}}, \theta_{\text{MLP}}} \left[ \sum_{(\mathbf{\mathcal{G}}^k, \mathbf{u}^k)\in \mathcal{D}} \mathcal{L}_{ce}(\Phi(\mathcal{G}^k),\mb{u}^k) \right]
\end{equation*}
For the training, the dataset is divided into a training set (90\%) and a validation set (10\%). The training is conducted with the Adam optimizer and the learning rate is scheduled to decay from $10^{-4}$ to $10^{-6}$ within 20 epochs with cosine annealing. Finally, we trained with batch size 16 and weight decay equal $10^{-5}$. 

\subsection{Network Architecture}
\label{sec:architecture}
The architecture features a single-layer Graph Block, described in Section \ref{subsection:Graph_Block}, to provide the back bone of the distributed coordination scheme. We parametrize the learnable, node update function $\phi^v(\cdot)$, illustrated in Fig.\ref{fig:Graph Block} as follows: Initially, a ResNet module implements feature extraction from the 4 channels and comprises of 3 sequential 2-layer (\textit{Conv2d-BatchNorm2d-ReLU}) Residual Blocks with skip connections, followed by an adaptive average-pooling module. The latter provides the network with the flexibility to process environments of variable size by projecting to a fixed representation. The final layer is an MLP with parameters $\theta_{\text{MLP}}$, that translates this representation into probabilities per control action. 


\section{Experiments}
\label{subsec:experiment_1}
In this Section, we provide simulations to demonstrate the efficacy of our method. All case studies have been implemented using Python3 on a computer with a 3.2 GHz, 8-Core, Intel Core i11-800H CPU, 64GB RAM and an Nvidia RTX 3080Ti GPU, 16GB RAM. The experiments were conducted assuming that the robots can take noisy measurements with a camera that provides the $xy$-coordinates of any visible object within its sensing range, expressed in the robot's frame, where the sensing radius is selected to be equal to $2m$ and the noise covariance matrix $\mathbf{R}_{i,t}= (0.05 \odot \mb{x}_t)^2\mathbf{I}$. We assume first-order dynamics robots that select any action from the set $\mathcal{U} = \{\textit{left}, \textit{right}, \textit{up}, \textit{down} \}$ by $0.5m$ and the DKF parameters are chosen as $\kappa_{ii} = 0.75$ and $\kappa_{ij} = 0.25/\vert \mathcal{N}_i \vert$. We test our algorithm on the applications of target localization and tracking, where the hidden state $\mathbf{x}_t$ collects the positions of all targets at time $t$, i.e $\mathbf{x}_{t}=[\mathbf{x}_{1,t}^T,\dots,\mathbf{x}_{M,t}^T]^T$, where $\mathbf{x}_{k,t}$ is the position of target $k$ at time $t$ and $M$ is the total number of targets. In this setting, we require to drop the uncertainty for all targets by at least one robot, i.e. $\det\Sigma_i^k \leq \epsilon$. To achieve collision avoidance when an invalid action is given, the robot randomly selects a collision-free action from $\mathcal{U}$, otherwise stays idle. A more sophisticated collision avoidance method \cite{long2018towards, alonso2013optimal, alonso2018cooperative} could be used instead. Unless stated differently, we assume a fully connected graph. It is important to stress that for all the experiments below we use the same network, trained as described in \ref{sec:dataset_training} and without any fine-tuning per task.

\subsection{Scalability and Generalization}
We are interested in examining how our network, trained on $20m \times 20m$ ($200 \times 200$ occupancy grid maps) with $N=10$ robots and a hidden state $\mathbf{x}_t \in \mathbb{R}^{20}$ would perform on previously unseen $20m \times 20m$ environments and varying number of robots $N$ and targets $M$. For this experiment, we run the Expert (C-SB) for each of the following instances $N\times M=[10\times10, 20\times20, 30\times30, 40\times40, 50\times50, 60\times60, 80\times 80]$ for $n=100$ random initializations (obstacles, robots and target positions) and compare our GNN approach with two baselines (a) a Random-Walker and (b) a Distributed Sampling-Based (Dec-SB) AIA algorithm \cite{tzes2021distributed}. We evaluate the algorithms similar to \cite{li2020graph} based on the following metrics: \newline
1) Flowtime Increase $\delta_{FT}= (FT - FT^*)/FT^*$, expresses the percentage change between the returned planning horizon $F$ and the expert's horizon $F^*$. For each of the configurations, we collect $n$ of the $\delta_{FT}$'s and return as the flowtime increase of the configuration the mean. \newline
2) $\text{Success Rate} = n_{\text{success}}/n$, where $n_{\text{success}}$ is the number of successful cases out of the $n$ cases of the expert. A case is considered successful when a solution is acquired with a planning horizon $F \leq 3F^*$. 

\begin{figure}[t]
\captionsetup{width=\linewidth,font=footnotesize}
    \begin{subfigure}[b]{0.237\textwidth}
        \centering
    \includegraphics[width=\textwidth]{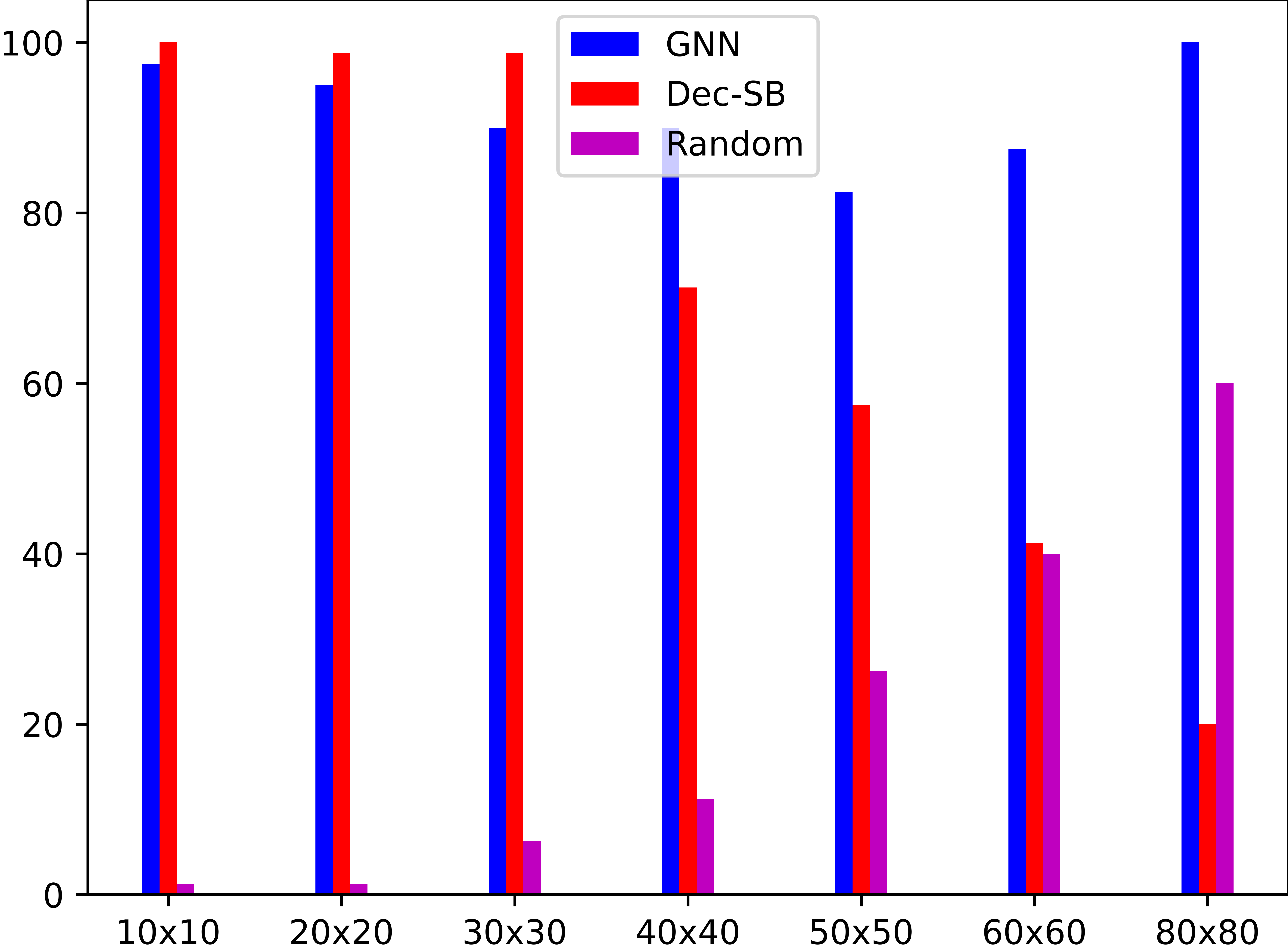}
    \caption{}
    \label{fig:success}
    \end{subfigure}
    \hfill
    \begin{subfigure}[b]{0.237\textwidth}
    \centering
    \includegraphics[width=\textwidth]{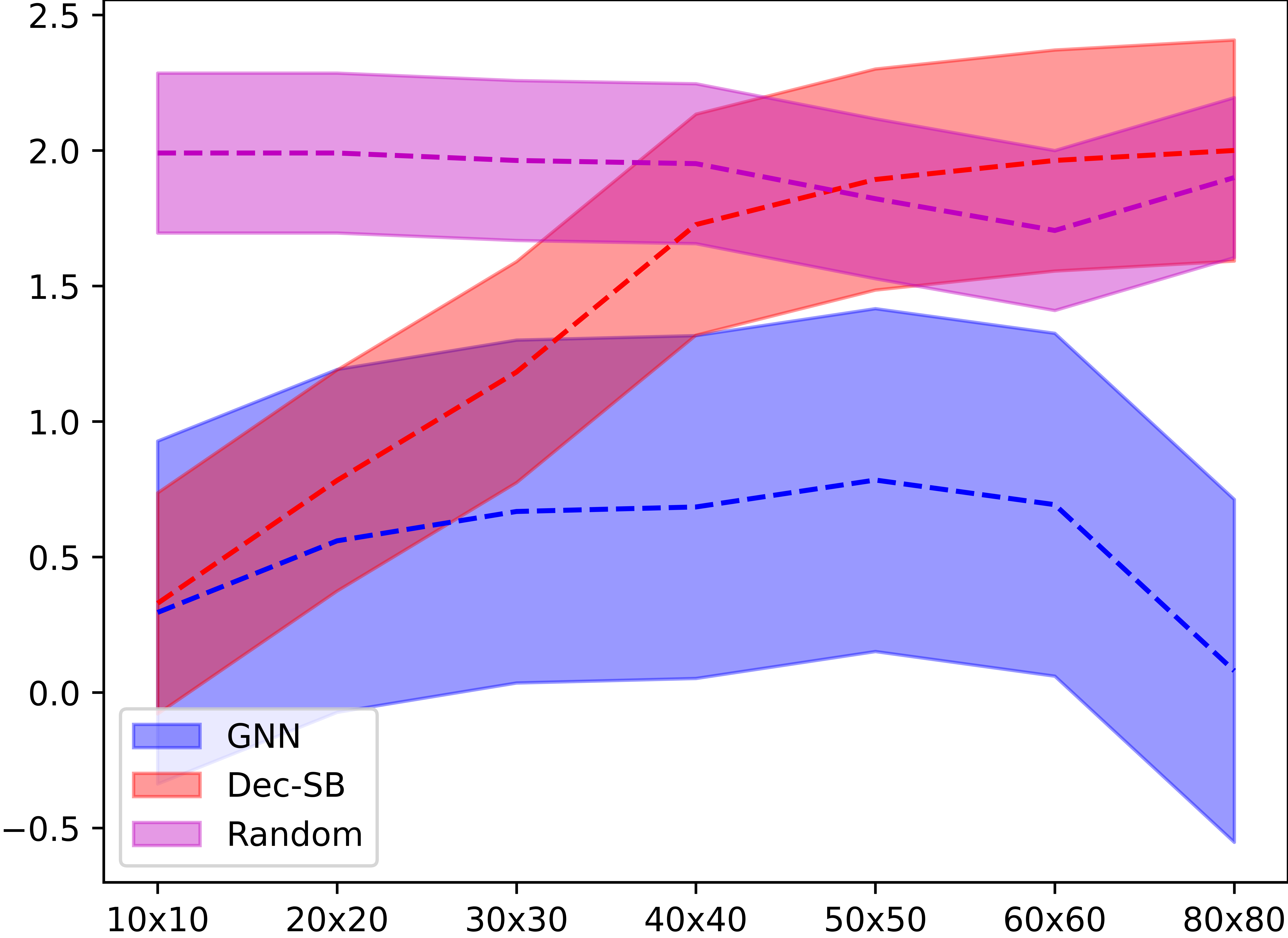}
    \caption{}
    \label{fig:flowtime}
    \end{subfigure}
    \setlength{\belowcaptionskip}{-18 pt}
    \caption{In this Figure we plot the success and flowtime increase for $N\times M$ robots and targets, where each problem instance is run for 100 random environments and initializations. We compare our GNN algorithm to two baselines a Decentralized Sampling-Based (Dec-SB) and a Random method}
    \label{fig:statistics}
\end{figure}

The results are presented in Fig. \ref{fig:statistics}. From Fig. \ref{fig:success}, we can see that the proposed GNN method performs equally well and occasionally better than the Dec-SB algorithm and always outperforms the Random Walker. The Random-Walker performs better on more condensed environments (robots randomly initialized closer to targets) while the Dec-SB could not frequently return solutions with planning horizons less than $3F^*$. Even in the case of condensed environments where we expect smaller planning horizons returned by the expert, our method manages to complete the tasks withing the horizon bounds. In Fig. \ref{fig:flowtime}, the mean of the Flow-Time increases of our method is systematically smaller than those of the baselines. We conclude that our network, trained only on environments of 10 targets and 10 robots is able to learn policies that generalize to novel environments with much larger graphs and hidden states. Moreover, mimicking a centralized expert is able to learn a distributed policy that is usually more effective than hand- designed distributed policies for the same problem.  

\subsection{Target Localization and Tracking}
\label{subsec:experiment_2}
\begin{figure}[t!]
\captionsetup{width=\linewidth,font=footnotesize}
\begin{subfigure}{0.24\textwidth}
\centering
\includegraphics[trim=0 0 0 1pt, clip,width=\textwidth]{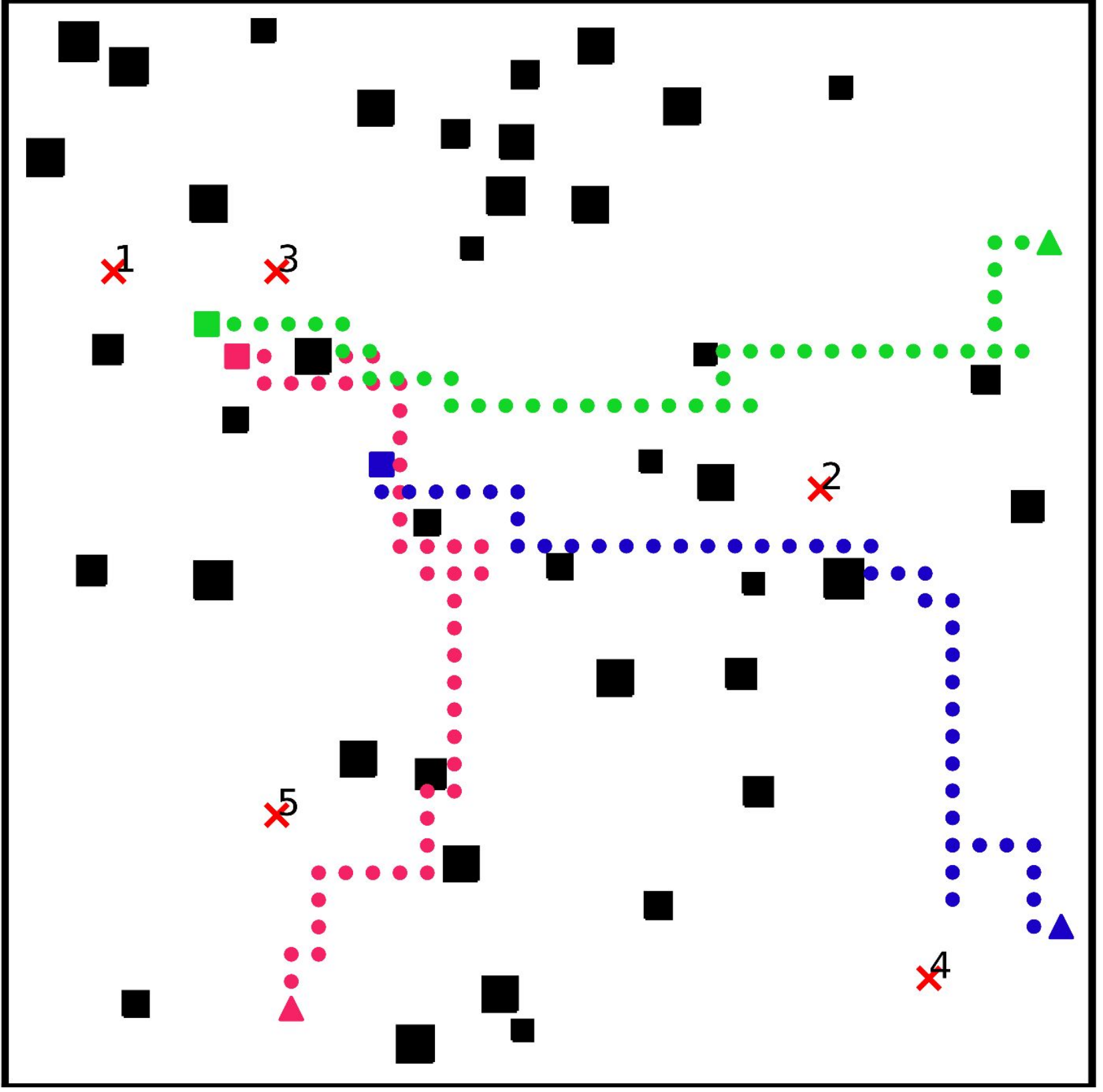}
\caption{}
\label{fig:static_traj}
\end{subfigure}
\hfill
\begin{subfigure}{0.24\textwidth}
\centering
\includegraphics[width=\textwidth]{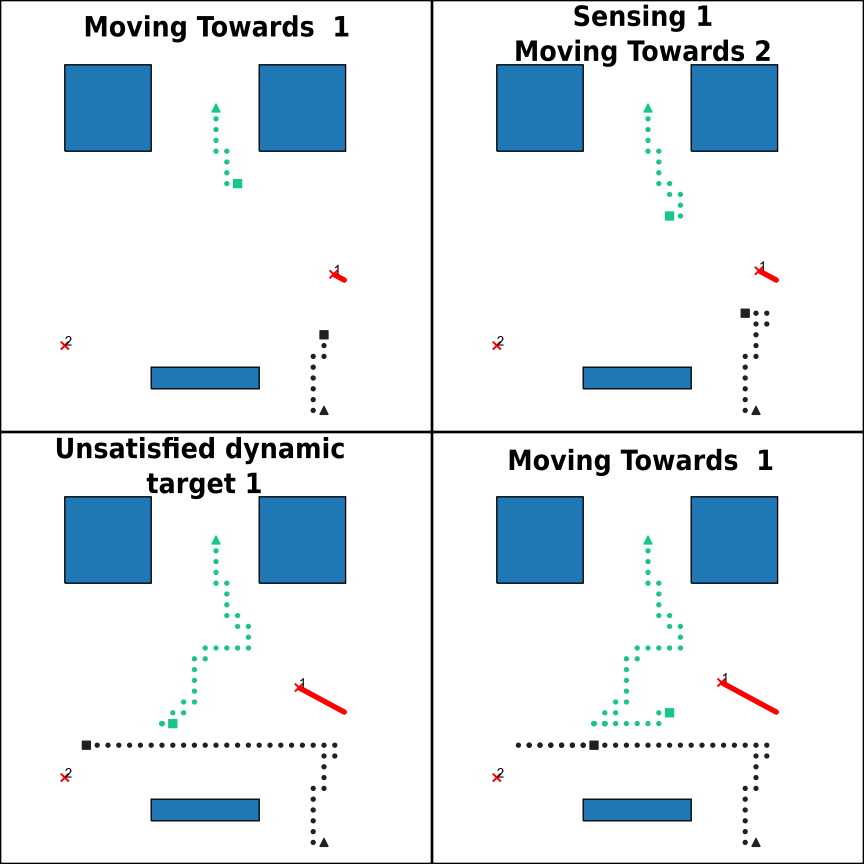}
\caption{}
\label{fig:dynamic_traj}
\end{subfigure}
\vfill
\begin{subfigure}{0.23\textwidth}
\centering
\includegraphics[width=\textwidth]{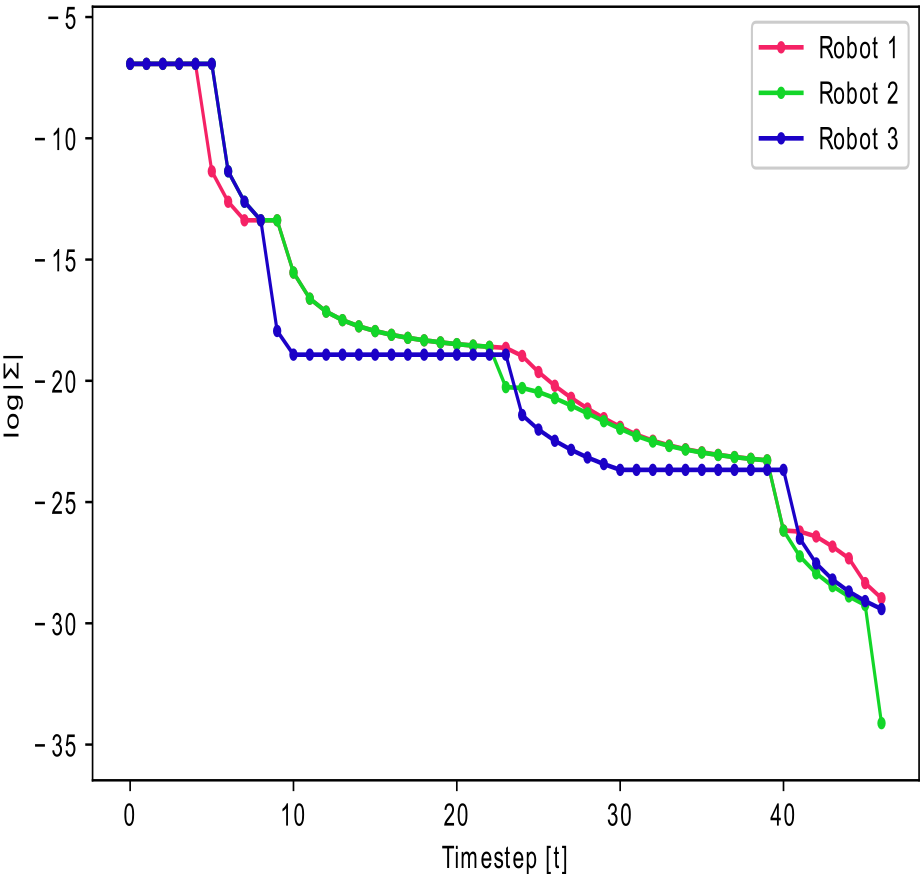}
\caption{}
\label{fig:static_sigmas}
\end{subfigure}
\hfill
\begin{subfigure}{0.23\textwidth}
    \centering
    \includegraphics[width=\textwidth]{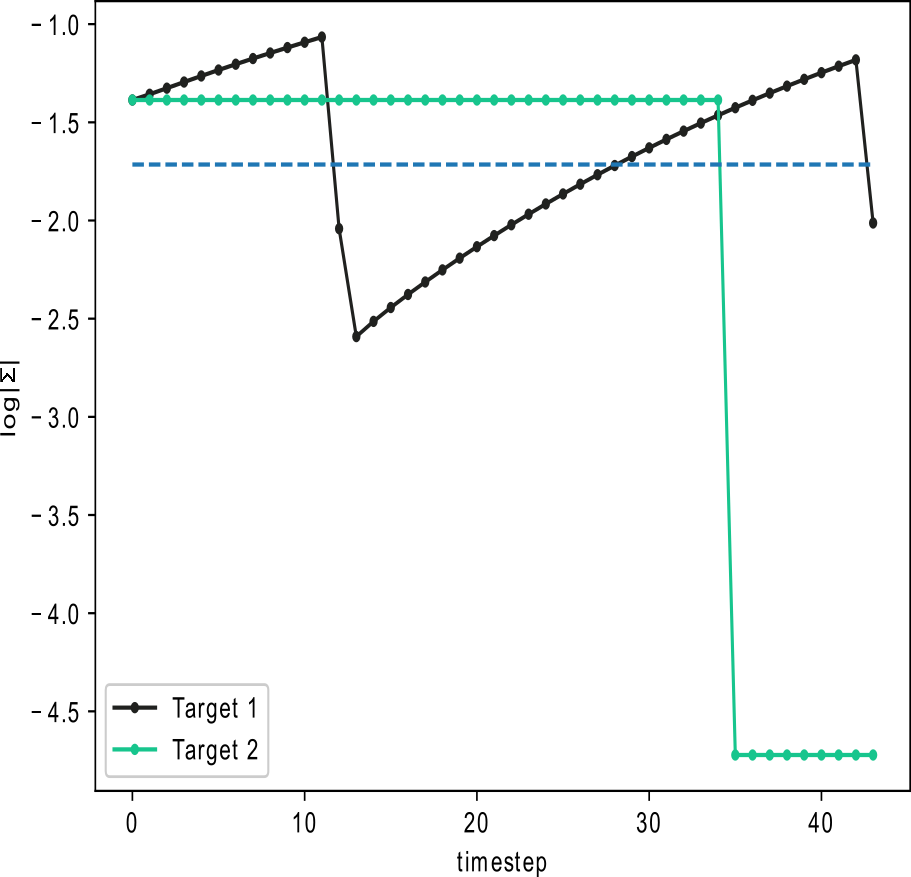}
    \caption{}
    \label{fig:dynamic_sigmas}
    \end{subfigure}
\setlength{\belowcaptionskip}{-18 pt}
\caption{This Figure illustrates qualitatively the efficacy of our scheme to the target localization and tracking problem. Fig. \ref{fig:static_traj} represents in different colors the executed trajectories of three robots tasked to localize 5 static targets. The evolution of their corresponding uncertainties is depicted in Fig. \ref{fig:static_sigmas}. In Fig. \ref{fig:dynamic_traj} we show spotlights of the solution with dynamic target and in Fig. \ref{fig:dynamic_sigmas} the evolution of the global covariance for each of the targets. In Fig. 4a,4b, the starting and ending position of the robots is illustrated as a triangle and a square respectively, while the targets as red $x$-markers.}
\label{}
\end{figure}
In this experiment, we are interested in validating our method qualitatively where the uncertainty threshold is set $\epsilon = 1.8e^{-2}$ for each of the targets. In the first scenario, we tested the robots reside on a randomly-generated $20m \times 20m$ environment cluttered with a dense number of obstacles. The robots $(N=5)$ are tasked to localize 3 static landmarks, while avoiding the  obstacles. From Fig. \ref{fig:static_traj}, we can see that the robots executed successfully their task and localized the targets. In Fig. \ref{fig:static_sigmas}, we draw the evolution of the determinants of the covariance matrices of the robots and certify that their uncertainties are decreasing with time. The robot depicted in green is the one that first achieves the desired ending condition. 

In the second scenario, we present how our algorithm responds to dynamic targets. Specifically, two robots are tasked to localize a static landmark and a dynamic target of known dynamics (see Section \ref{sec:problem_formulation}). In Fig. \ref{fig:dynamic_traj} we present spotlights of the online execution. From left-to-right, up-to-bottom the robots start moving towards target 1. Once they localize it, they start heading towards the next target. In the meantime, as the target moves, the increase in the uncertainty becomes evident on the robots' heatmap and thus return to re-localize it. The scope of Fig. \ref{fig:dynamic_sigmas} is to represent the global uncertainty for each of the targets defined at each timestep as the minimum of the determinants of the robots' covariance matrices. From the figure, it is clear that the robots successfully dropped both of the targets' uncertainty below the threshold, depicted as a dashed blue line. 

The online execution of our proposed scheme allows for an online update of the estimated location $\bm{\mu}_t$ of the dynamic target when sensed, in contrast to the centralized approach where the estimated locations remain the same. This flexibility makes our algorithm more robust to poor prior estimates $\hat{\mb{x}}_0$ and eliminates the need for replanning like in the expert's case (\cite[Section VI.C]{9811642}), since the updated mean will be immediately reflected in the heatmap.

\subsection{Robustness to Robot \& Communication failure}
\label{subsec:experiment_3}
\begin{figure}[t!]
\centering
\begin{subfigure}{0.3\textwidth}
\centering
\includegraphics[width=0.9\textwidth]{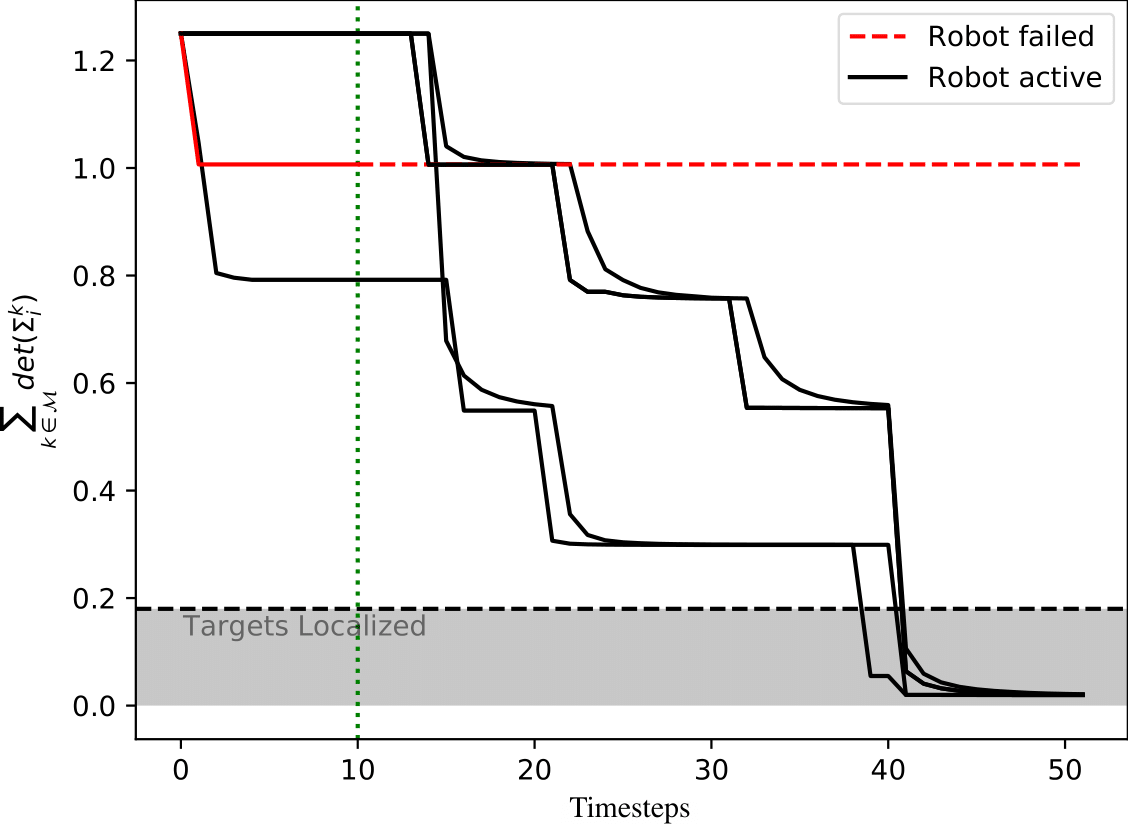}
\caption{}
\label{fig:resilience1}
\end{subfigure}
\hfill
\begin{subfigure}{0.15\textwidth}
\includegraphics[clip,trim=0.18cm 0.18cm 0 0 ,width=\textwidth]{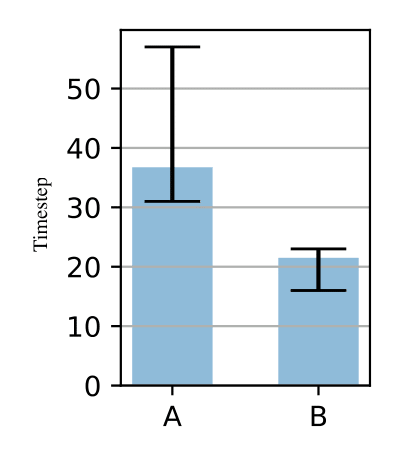}
\caption{}
\label{fig:time_increase}
\end{subfigure}
\vfill
\begin{subfigure}{0.15\textwidth}
\centering
\includegraphics[clip,trim = 1pt 0 21pt 0, width=\textwidth]{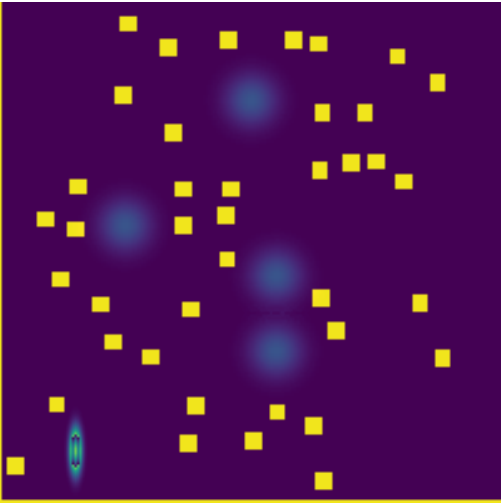}
\caption{}
\label{fig:t9_0}
\end{subfigure}
\begin{subfigure}{0.15\textwidth}
\centering
\includegraphics[clip,trim = 1pt 0 21pt 0, width=\textwidth]{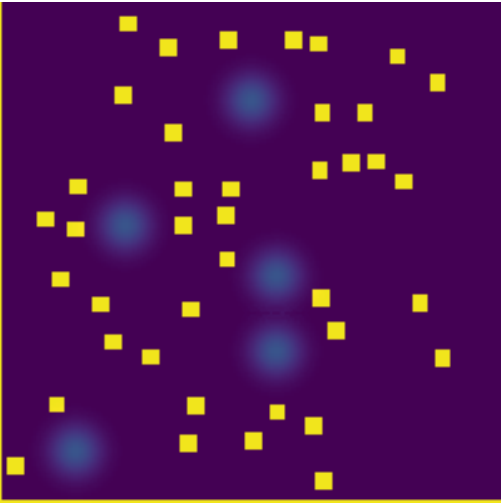}
\caption{}
\label{fig:t9_1}
\end{subfigure}
\hfill
\begin{subfigure}{0.15\textwidth}
\centering
\includegraphics[clip,trim = 1pt 0 21pt 0, width=\textwidth]{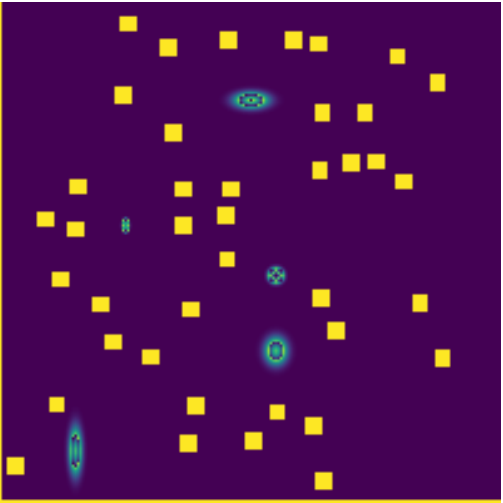}
\caption{}
\label{fig:t49_1}
\end{subfigure}
\setlength{\belowcaptionskip}{-18 pt}
\caption{Figure \ref{fig:resilience1} depicts the per-node sum of determinants over time, with black lines representing active robots, red the one that fails at $t=10$, and gray area the localization threshold. Figures \ref{fig:t9_0}, \ref{fig:t9_1} and \ref{fig:t49_1} illustrate the hidden state distribution at $t=9$ for robots $k$ and $2$ and at time $t=49$ for robot $2$ respectively. The average mission time for 50 tests is illustrated in Figure \ref{fig:time_increase}, where $\mathcal{A}$ refers to the case study with distance-defined communication graph, communication loss and agent failure and $\mathcal{B}$ the one with a static fully connected graph.}
\end{figure}
This experiment examines the adaptability of I-GBNet to robot failure in target localization missions. In a $20m \times 20m$ obstacle riddled environment we assign $N=5$ robots to localize $M=5$ targets. The graph connectivity varies over time owing to the mobility of the agents for a given communication range $r_{com} = 4m$. We test for possible loss of packets during communication by introducing edge deletion based on a Poisson distribution $(\lambda = 1)$. At time $t^{fail}=10$, a randomly selected agent $k$ exhibits catastrophic failure and is permanently isolated from the swarm. Our GNN-based approach demonstrates online robustness to alterations in the synthesis of the swarm. Despite robot $k$ failing, depriving the swarm of any information individually acquired through its sensing capabilities (see Figures \ref{fig:t9_0} and \ref{fig:t9_1}; bottom-left target has been localized only from agent $k$ that is about to fail without propagating that information), the swarm re-deploys other robots to replace any acquired measurements from the failed robot not transmitted in the graph prior to failure, thus managing to localize all targets (see Figure \ref{fig:t49_1}; the swarm has managed to localize the bottom-left target). All the active robots reach a precision consensus regarding the determinants of the covariance matrices, i.e. $\sum_{m\in \mathcal{M}}\det\Sigma^m_{i,t}$, below the required threshold for localization (see Fig. \ref{fig:resilience1}). However, robot or communications failure appear to have prolonged the average mission time by $71.51\%$ (see Fig.\ref{fig:time_increase}), as information propagation in the network is constantly impeded.

\section{Conclusions}
In this paper, we reduce the multi-robot Active Information Acquisition problem into a learning framework over graphs and introduce I-GBNet, an AIA-inspired GNN, that offers distributed decision-making, imitating a centralized expert. In contrast to other works, our method can deal with dynamic phenomena and is robust to changes in the communication graph. Experiments demonstrated that despite being trained on small graphs, our network successfully generalizes to previously unseen environments/robot configurations and scales well. In the future, we shall investigate on more applications of Active Information Acquisition, like occupancy mapping and we shall explore the applicability of attention layers to further encourage scalability and time-efficiency. 

\bibliographystyle{IEEEtran}
\bibliography{bibliography_mariliza,bibliography_nikos}

\end{document}